%% file: main.tex
\documentclass[conference]{IEEEtran}
\IEEEoverridecommandlockouts
\usepackage{cite}
\usepackage{amsmath,amssymb,amsfonts}
\usepackage{algorithmic}
\usepackage{graphicx}
\usepackage{textcomp}
\usepackage{xcolor}
\usepackage{booktabs}
\usepackage{multirow} 

\usepackage{subcaption} 
\usepackage{siunitx}  
\usepackage{hyperref}
\usepackage{svg}
\def\BibTeX{{\rm B\kern-.05em{\sc i\kern-.025em b}\kern-.08em
    T\kern-.1667em\lower.7ex\hbox{E}\kern-.125emX}}
\begin{document}

\title{Early Alzheimer’s Disease Detection from Retinal OCT Images: A UK Biobank Study\\
\thanks{This study was supported by Scientific and Technological Research Council of Turkey (TUBITAK) under the Grant Number 122E509. The authors thank to TUBITAK for their supports.}
}

\author{\IEEEauthorblockN{1\textsuperscript{st} Yasemin Turkan }
\IEEEauthorblockA{\textit{Department of Computer Engineering} \\
\textit{Isik University}\\
Istanbul, Turkey \\
}
\and
\IEEEauthorblockN{2\textsuperscript{nd} F. Boray Tek }
\IEEEauthorblockA{\textit{Department of Artificial Intelligence and Data Engineering} \\
\textit{Istanbul Technical University}\\
Istanbul, Turkey \\
}
\and
\IEEEauthorblockN{3\textsuperscript{rd} M. Serdar Nazlı }
\IEEEauthorblockA{\textit{Department of Artificial Intelligence and Data Engineering} \\
\textit{Istanbul Technical University}\\
Istanbul, Turkey \\
}
\and
\IEEEauthorblockN{4\textsuperscript{th} Öykü Eren }
\IEEEauthorblockA{\textit{Department of Artificial Intelligence and Data Engineering} \\
\textit{Istanbul Technical University}\\
Istanbul, Turkey \\
}

}

\maketitle

\begin{abstract}
 Alterations in retinal layer thickness, measurable using Optical Coherence Tomography (OCT), have been associated with neurodegenerative diseases such as Alzheimer's disease (AD). While previous studies have mainly focused on segmented layer thickness measurements, this study explored the direct classification of OCT B-scan images for the early detection of AD. To our knowledge, this is the first application of deep learning to raw OCT B-scans for AD prediction in the literature. Unlike conventional medical image classification tasks, early detection is more challenging than diagnosis because imaging precedes clinical diagnosis by several years. We fine-tuned and evaluated multiple pretrained models, including ImageNet-based networks and the OCT-specific RETFound transformer, using subject-level cross-validation datasets matched for age, sex, and imaging instances from the UK Biobank cohort. To reduce overfitting in this small, high-dimensional dataset, both standard and OCT-specific augmentation techniques were applied, along with a year-weighted loss function that prioritized cases diagnosed within four years of imaging. ResNet-34 produced the most stable results, achieving an AUC of 0.62 in the 4-year cohort. Although below the threshold for clinical application, our explainability analyses confirmed localized structural differences in the central macular subfield between the AD and control groups. These findings provide a baseline for OCT-based AD prediction, highlight the challenges of detecting subtle retinal biomarkers years before AD diagnosis, and point to the need for larger datasets and multimodal approaches.
\end{abstract}


\input{intro}
\input{method}

\input{results}

\input{discussion}

\input{conclusion}

\section*{Acknowledgements}
This study was conducted using the UK Biobank Resource under Application Number 82266. Computational resources were provided by the Turkish National High-Performance Computing Center (UHeM, Project Number 1017802024). This study was also supported by the Scientific and Technological Research Council of Turkey (TÜBİTAK) under Grant Number 122E509.

\bibliographystyle{IEEEtran}
\bibliography{ubmk} 

\end{document}

%% file: intro.tex
\section{Introduction}
Alzheimer's disease (AD) is an irreversible and progressive brain disorder characterized by a decline in cognitive function and is the most prevalent type of dementia. Currently, there is no known cure, but it is marked by a significant reduction in brain size (neurodegeneration) caused by the accumulation of proteins (amyloid-beta and tau) in neurons\cite{WHO2019}. As the retina and brain originate from the same neural tube, the eyes are often regarded as extensions of the brain \cite {Grzybowski2020}. Postmortem studies have shown that amyloid-beta and tau proteins accumulate in the retinas of individuals with AD\cite{London2013}. 

High-resolution visual imaging technologies, such as optical coherence tomography (OCT),  have recently been proposed to examine the structural and vascular changes in the retinas of patients with AD.  In contrast to current gold-standard diagnostic tools, such as positron emission tomography (PET), cerebrospinal fluid (CSF) analysis, and genetic screening, OCT offers a non-invasive, rapid, and cost-efficient imaging modality that is already widely available in clinical ophthalmology settings. Incorporating such retinal imaging biomarkers into clinical workflows could help identify at-risk individuals for further neurological testing, thus enabling earlier intervention strategies and improving patient outcomes in the future.

Existing studies have shown statistical associations between retinal layer thickness and AD. However, to our knowledge, this is the first study to apply deep learning to raw OCT B-scan images from the UK Biobank for early AD prediction. This represents a significant methodological advance over prior studies, which primarily relied on retinal fundus images \cite{Tian2021, Yousefzadeh2024} or quantitative summary measurements relying on a layer segmentation procedure \cite{Heide2024} rather than on raw, high-resolution structural scans.

Our study introduces an end-to-end deep learning framework that includes anatomically guided pre-processing, multichannel input design, and rigorous cross-validation tailored for OCT data. We evaluate multiple state-of-the-art models, including CNNs and vision transformers, to assess the feasibility of learning subtle retinal biomarkers from small, temporally distant datasets.

The dataset,  model, and methodology are explained in Section \ref{sec:Methodology}. Section \ref{sec:Results} describes the experiments and presents the results of this study. Our discussion and conclusions are presented in Sections \ref{sec:Discussion} and \ref{sec:Conclusion}, respectively.

%% file: method.tex
\section{Methods}
\label {sec:Methodology}
We adopted a standardized deep learning pipeline in alignment with the Deep Learning Flow framework tailored for AD/MCI diagnosis using OCT-OCTA \cite{Turkan2024}. This framework emphasizes the best practices for the curation of retinal OCT datasets, model training, and evaluation protocols. This modular structure emphasizes the reproducibility and fairness of the framework in developing deep learning models for ophthalmic imaging.

\subsection {Dataset Curation}
\subsubsection*{Inclusion and Exclusion}

The UK Biobank database includes 502,386 participants, of whom 85,704 underwent OCT tomography scans at instances 0 and 1 using the same OCT device, Topcon 3D OCT 1000 Mk 2 \cite{Chua2019}. Until July 2023, 3,955 patients were diagnosed with Alzheimer's disease (AD). However, only 539 patients with AD underwent corresponding OCT scans.  Low-quality scans were excluded based on the  criteria used by Patel et al.\cite {Patel2016, Heide2024}. After applying the exclusion criteria, 43,934 participants remained in the dataset, with only 223 of them diagnosed with Alzheimer's disease (AD). The dataset included patients who were diagnosed over a 12-year period. Due to the slow progression of AD, more significant changes are typically observed in patients at least three years prior to diagnosis. Since only 9 patients with AD had reached the three-year mark in our dataset, a four-year threshold was selected to ensure a larger and more representative sample size (19 patients with AD). Table \ref{table:cohort_summary} summarizes the number of cohorts in the full dataset versus AD at each data-exclusion iteration. 

Both eyes of participants in the AD cohort were included in the study if they were eligible for the study. However, to prevent data leakage during model development, both eyes from a single patient were assigned exclusively to the training or validation set. Age, sex, and instance matched individuals from the non-AD group were randomly selected to form a balanced control group. 

\begin{table}
\centering
\caption{Summary of Cohorts with Dementia and AD Counts and Percentages}
\begin{tabular}{lcccccc}
\toprule
\textbf{Dataset}&\textbf{Cohort} & \multicolumn{2}{c}{\textbf{AD}} \\ 
\cmidrule(lr){3-4} 
& \textbf{N} & \textbf{N} & \textbf{\%} \\ 
\midrule
UK Biobank cohort & 502,386  & 3955 & 0.79\% \\
OCT-Scanned cohorts & 85,704  & 539 & 0.65\% \\
High Quality Cohorts with OCT scans & 43,934  & 223 & 0.51\%\\
4-Year dataset cohorts (Dataset 1) & 49  & 19 & 39\%\\
4-Year dataset scans (Dataset 1) & 58 & 28 & 48\%\\
4-Year dataset cohorts (Dataset 2)  & 45  & 19 & 42\%\\
4-Year dataset scans (Dataset 2) & 55 & 28 & 51\%\\
\bottomrule
\end{tabular}
\label{table:cohort_summary}
\end{table}

\subsubsection*{Image Pre-process and Feature Extraction} 

\begin{figure}
\centering
\includegraphics[width=0.47\textwidth]{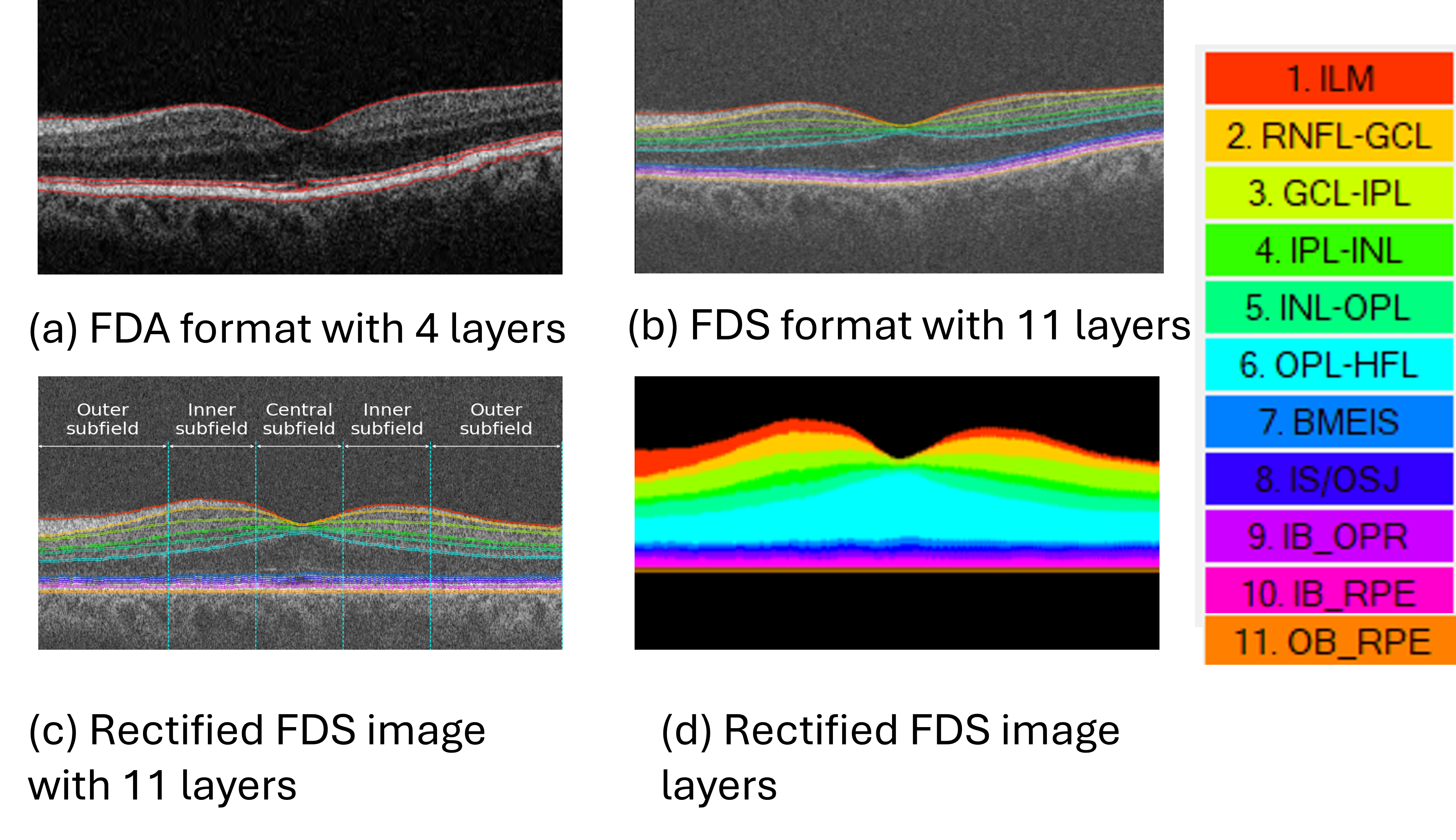}
\caption{Overview of the preprocessing pipeline and retinal layer annotations in OCT B-scans.
(a) Original grayscale OCT scan with inner and outer retinal boundaries. (b) Retinal layer segmentation with 11 color-coded contours representing the anatomical boundaries. (c) Alternate view of the same segmented B-scan for visualization consistency. (d) Pixel-wise retinal layer mask used as input for the second channel of the model. The legend on the right maps each color to a specific retinal layer, from the inner limiting membrane (ILM) to the outer boundary of the retinal pigment epithelium (OB\_RPE). This multichannel representation encodes both intensity and anatomical structure, providing richer input for deep learning models. \textit{Images used with permission from the UK Biobank under Application Number 82266.}}
\label{fig:ImagePreprocess}       
\end{figure}

In the UK Biobank,  FDS bulk files contain the entire, 16-bit raw OCT volume for both the left and right eyes, providing the complete dynamic range and unprocessed scan data necessary for custom image processing or resegmentation. In contrast, the FDA bulk files are an 8-bit downsampled representation derived directly from  FDS volumes; they include four segmentation layers but do not permit recovery of the original 16-bit voxel intensities. The differences between the FDA and FDS image formats are shown in Fig. \ref{fig:ImagePreprocess} (a) and (b). In this study, we used the OCTExplorer software, which utilizes the Iowa Reference Algorithms (Retinal Image Analysis Lab, Iowa Institute for Biomedical Imaging, Iowa City, IA) \cite{Garvin2009, Abramoff2010} for the segmentation of the fds files (Fig. \ref{fig:ImagePreprocess} (c)).

This study focused on ImageNet-pretrained 2D models. Therefore, the middle slice (B-scans) of the raw image volumes (128 × 512 × 650)  was selected as the input. The OCT image was rectified using the bottom contour, as shown in Fig. \ref{fig:ImagePreprocess} (c).  The areas above the top contour and below the bottom contour were removed to eliminate noise at the top and bottom of the scans. An additional layer-masked image was generated (Fig. \ref{fig:ImagePreprocess} (d) ). Finally, all pre-processed images of size 512 × 650 were cropped to 512 × 512. 

\begin{figure}
\centering
\includegraphics[width=0.5\textwidth]{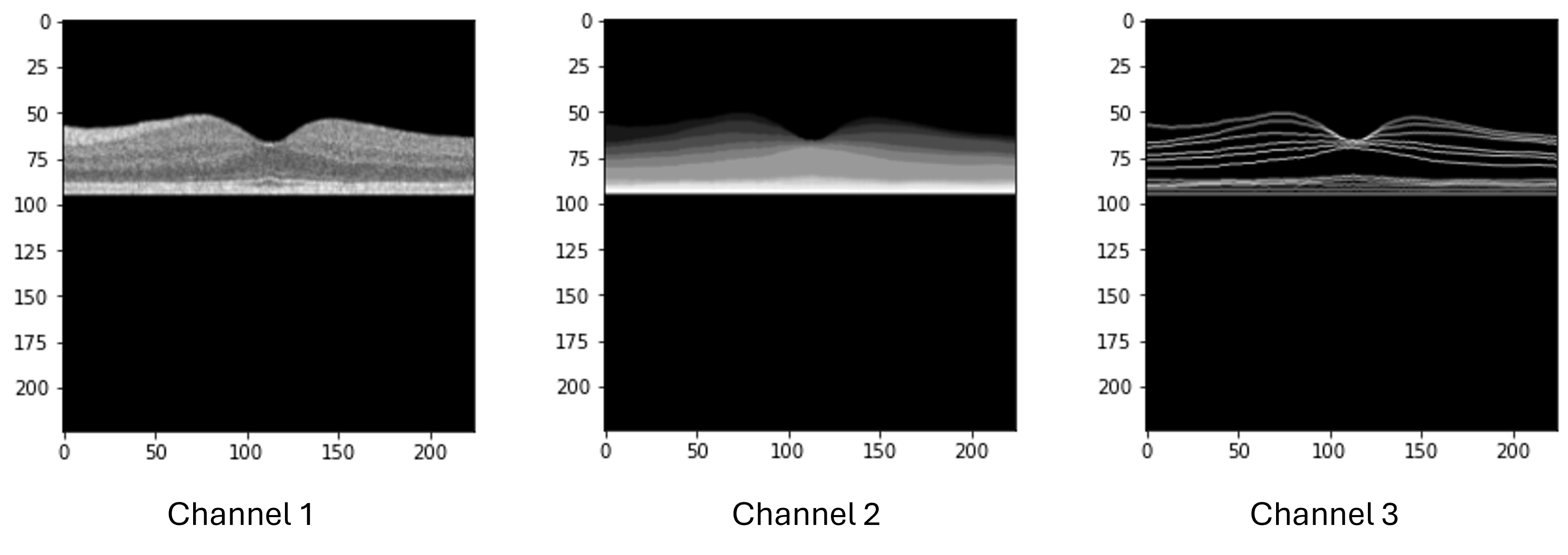}
\caption{Composite RGB representation of a single OCT B-scan used as model input. (a) First channel: original grayscale OCT image. (b) The second channel:  the layer-masked version, where the retinal layers are selectively enhanced to highlight the structural features. (c) Third channel:  binary retinal layer contours providing anatomical boundary information. \textit{Images used with permission from the UK Biobank under Application Number 82266.}}
\label{fig:input_img}       
\end{figure}

To adapt the single-channel OCT images to the three-channel input requirements of the pretrained models, we constructed a composite 3-channel representation for each sample, as shown in Fig. \ref{fig:input_img}. Specifically, the original grayscale OCT B-scan was assigned to the first channel of the network. The second channel contained a layer-masked version of the image, where each retinal layer was selectively enhanced to highlight the structural regions of interest \cite{Eren2024}. The third channel consisted of a binary image encoding the retinal layer contours, providing explicit anatomical boundaries as auxiliary input. This multichannel representation was designed to enrich the input with both intensity-based and structural information. This facilitates improved feature extraction in downstream convolutional-based architectures.

\subsection{Training}
\subsubsection*{Augmentation} 
Deep learning models are prone to overfitting, particularly when they are trained on small datasets. To mitigate this issue and improve generalization, we applied extensive augmentation strategies to artificially expand the training set and introduce greater variability, thereby enhancing the robustness of the model and reducing the risk of overfitting. During training, the augmentation techniques were randomly selected and applied one at a time. Custom image augmentations, such as flipping and affine transformations (excluding shear and rotation in rectified images because of the nature of these images), were applied to all channels. In addition, \textbf{OCT-specific augmentations} such as occlusion, contrast adjustment, artificial vascular patterns, and noise addition were employed only in the first channel of the model.

\subsubsection*{Training Models and Parameters} 

To investigate the effectiveness of deep learning for OCT-based classification, we employed a pretrained convolutional neural network (CNN) and transformer architecture. Specifically, we used ResNet (depths of 18,34,50, and 101) and VGG (depths of 8 and 11). 

In addition to these standard architectures, we also tested a domain-specific RETFound \cite{Zhou2023}, a vision transformer pretrained on a large dataset (over a million) of OCT and fundus images. RETFound has demonstrated strong performance in retinal imaging tasks \cite{Nazli2024} and was therefore included for comparative evaluation in our Alzheimer’s disease classification framework. We implemented two distinct training strategies: \textbf{RETFound‑S}, trained using three identical mid‑scan OCT images as input, and \textbf{RETFound‑C},  trained end‑to‑end using the composite 3-channel representation described in Fig. ~\ref{fig:input_img}.

We also developed and trained a custom convolutional model to serve as a baseline and explored architectural variations tailored to the specific characteristics of OCT data.

All models were trained using a batch size of 4, which was selected to maintain stable gradient updates while accommodating the constraints of small dataset size. We experimented with a range of learning rates (0.001, 0.0001, and 0.000027) to explore the optimal convergence behavior across different model architectures. The AdamW optimizer was used for optimization in all experiments. We employed extensive image augmentation techniques (as detailed in the previous section) to address the risk of overfitting.

To adapt the pretrained backbones for binary classification under low-data conditions, we appended a lightweight classification head consisting of a fully connected layer (from the feature dimension to 64 units), Layer Normalization, followed by a ReLU activation, dropout (p = 0.4), and a final linear layer projecting to two output classes. This minimal regularization structure was selected to balance the expressiveness and overfitting control.

We employed a year-weighted loss function\cite{Nazli2024}, assigning higher importance to samples temporally closer to the clinical diagnosis of Alzheimer’s disease to recognize disease progression dynamics. 

Due to the extensive data augmentation and regularization techniques applied to mitigate overfitting, the training and validation accuracy exhibited considerable fluctuations, making it difficult to identify an appropriate stopping epoch for the model. To stabilize the training and improve convergence, we trained the models for 100 epochs and then applied Stochastic Weight Averaging (SWA) \cite{Izmailov2019} starting from epoch 80.

The source code supporting the findings of this study will be released upon publication in the GitHub repository: https://github.com/OCTALZ-Project/UKBiobankAD

\subsection {Validation}
\subsubsection*{Comparing Results}
We employed a nested cross-validation strategy \cite{Zhong2023} for a robust and unbiased performance estimation. The data were split into five outer folds, each of which served as a held-out test set. Within the remaining training data, three-fold inner cross-validation was used to tune the hyperparameters. The selected model was evaluated using the corresponding test fold. The predictions from the five outer folds were pooled to compute a single AUC per run. This procedure was repeated five times with different random splits, yielding five AUC values per model (5 runs × 5 outer folds × 3 inner folds).

We applied this procedure separately to ResNet (18, 34, 50, and 101), VGG (8 and 11), RetFound (S and C), and the CNN baseline models. For each architecture, we calculated the mean AUC across five runs and selected the top-performing variant within each family. These best variants were then compared using a calibrated paired t-test \cite{bouckaert2003}, with adjusted degrees of freedom to account for dependence on repeated cross-validation.

\subsubsection* {Explainability} 

We chose the Grad-CAM method  to explain our best-performing model.  Instead of using the standard image overlay format, we used 10 OCT-specific layers and the central subfield of the macula.  We applied a 0.8 threshold to the Grad-CAM results to generate a highly focused explainability image. The contours of the OCT scan were overlaid on this image.  To evaluate the class-level performance, we used the Intersection Over Union (IoU), Dice Score, and Filling Ratio on each layer, as detailed in the study by Nazli et al.\cite{Nazli2024}. Finally, we added the center subfield region of the OCT b-scans,  where the overlaid GradCAM results were highlighted. 

\subsubsection* {Running with another sample set} 
We generated another training dataset with different age, sex, and instance-matching CNs while keeping the AD population the same. For the best-performing model, the same experiments were performed to determine whether the results were consistent with different datasets.

\subsubsection* {Ablation Studies} 
The model trained with only one of the channels was replicated for all three channels to determine the impact of the segmentation and contour information on the training results.

%% file: results.tex
\section{Results}
\label{sec:Results}

We evaluated multiple deep learning models using nested cross-validation, 5 outer folds, and 3 inner folds to assess the classification performance in predicting Alzheimer’s disease (AD) using retinal OCT scans acquired 4 years before diagnosis. Table \ref{tab:acc_results} lists the top AUC results obtained for each model. 

The highest performing model was ResNet-34, which achieved a mean AUC of 
$\mathbf{0.624 \pm 0.060}$ across five nested cross-validation runs. 
This model was used as a reference point for pairwise comparisons with alternative architectures. 
VGG-11 (mAUC $=0.581 \pm 0.017$) and RETFound-C (mAUC $=0.540 \pm 0.037$) 
showed reduced performance relative to ResNet-34, but the differences were not statistically 
significant (corrected paired $t$-tests $p=0.1458$ and $p=0.0845$). 
The custom CNN model (mAUC $=0.519 \pm 0.026$) also performed worse than ResNet-34, 
with significance under the standard $t$-test ($p=0.0266$), but not after correction ($p=0.0364$). 
In contrast, RETFound-S (mAUC $=0.459 \pm 0.068$) was significantly worse than ResNet-34 
under both tests (standard $p=0.0043$; corrected $p=0.0202$). 

We further validated the ResNet architecture on an independent sample set (Dataset 2), 
which achieved a very similar performance (mAUC $=0.652 \pm 0.058$), 
This supports the robustness of the model. 

Finally, ablation experiments were conducted by rerunning ResNet-34 with reduced feature inputs. 
When trained with masked images only (ResNet\textsuperscript{1}, mAUC $=0.452 \pm 0.042$), 
OCT images replicated to three channels (ResNet\textsuperscript{2}, mAUC $=0.561 \pm 0.061$), 
or layer contour inputs (ResNet\textsuperscript{3}, mAUC $=0.529 \pm 0.071$), 
The performance decreased significantly compared to that of the full multichannel representation. 
These results confirm that combining raw OCT, masked, and contour information 
provides the strongest feature representation for early AD prediction.

%%%%%%%%% <<SERDAR BEGINS>> %%%%%%%%%%%%%%%
\begin{table*}[ht]
\centering
\caption{Model accuracies on 4-Year dataset}
\label{tab:acc_results}
\begin{tabular}{l c c c c c c c c}
\textbf{Model} &  \textbf{mAUC} &  f1-score & Precision & Sensitivity & Specificity  & \textbf{t-test} & \textbf{corrected t-test} \\ 
 &&&&  & & \textbf{p Value}   & \textbf{p Value} \\ 
\hline

\textbf{ResNet}    &      $\mathbf{0.624} \pm \mathbf{0.060}$ &$\mathbf{0.552} \pm \mathbf{0.135}$ & $\mathbf{0.583} \pm \mathbf{0.086}$ &  $\mathbf{0.680} \pm \mathbf{0.018}$ & $0.486 \pm 0.159$ &   \\
VGG              & $0.581 \pm 0.017$ & $0.527 \pm 0.064$ & $0.568 \pm 0.055$ & $0.633 \pm 0.062$  &  $0.500 \pm 0.076$  & 0.1354  &0.1458  \\ 
RETFound-C    & $ 0.540\pm 0.037$  & $0.524 \pm 0.061$  & $0.557 \pm 0.039$ &$0.607 \pm 0.037$  & $\mathbf{0.507} \pm \mathbf{0.081}$ &0.0728  &  0.0845    \\ 
Custom CNN model     & $ 0.519\pm 0.026$ &  $0.490 \pm 0.039$ &$0.553 \pm 0.028$ & $0.600 \pm 0.041$& $0.464 \pm 0.051$ & *0.0266  &*0.0364 \\ 
RETFound-S    & $ 0.459\pm 0.068$ &  $0.468 \pm 0.072$ &  $0.461 \pm 0.046$  & $0.446 \pm 0.122$ &$0.479 \pm 0.117$ &*0.0043 & *0.0202    \\ 
\hline
ResNet\textsuperscript{1}         & $0.452\pm 0.042$ &$0.424 \pm 0.018$ & $0.463 \pm 0.021$ &$0.520 \pm 0.038$ & $0.407 \pm 0.020$ & *0.0133 &  *0.0208 \\
ResNet\textsuperscript{2}         & $0.561 \pm 0.061 $  & $0.523 \pm 0.095$& $0.531 \pm 0.090$& $0.527 \pm 0.086$ & $0.536 \pm 0.107$  & *0.0464 &  0.0577 \\
ResNet\textsuperscript{3}         & $ 0.529\pm 0.071$  & $0.498 \pm 0.091$& $0.517 \pm 0.081$ &  $0.533 \pm 0.071$ & $0.500 \pm 0.104$ & 0.0643& 0.0759  \\
\hline
ResNet\textsuperscript{4}         & $\mathbf{0.652} \pm \mathbf{0.058}$ & $\mathbf{0.613} \pm \mathbf{0.037}$& $\mathbf{0.604} \pm \mathbf{0.051}$ &$0.593 \pm 0.094$ &$\mathbf{0.614} \pm \mathbf{0.030}$  & 0.6815 &  0.6773 \\
\end{tabular}
    % Use this minipage for a detailed note under the table
    \begin{minipage}{\textwidth}
        \small % Make the text a bit smaller
        \textit{Note:} ResNet\textsuperscript{1}: the ablation test run with masked images replicated to 3 channels. ResNet\textsuperscript{2}: ablation test run with OCT images replicated into three channels. ResNet\textsuperscript{3}: The ablation test was performed with layer contours replicated into three channels. ResNet\textsuperscript{4}: the ResNet model trained with the dataset 2. Corrected $p$-values were computed using a calibrated paired $t$-test to account for dependence  induced by repeated cross-validation~\cite{bouckaert2003}.
    \end{minipage}

\end{table*}

\begin{figure}
    \centering

    % --- Subfigure (a): The Top 5% Regions ---
    \begin{subfigure}[b]{0.95\columnwidth} % Use \columnwidth for single-column figures
        \centering
        \includegraphics[width=\textwidth]{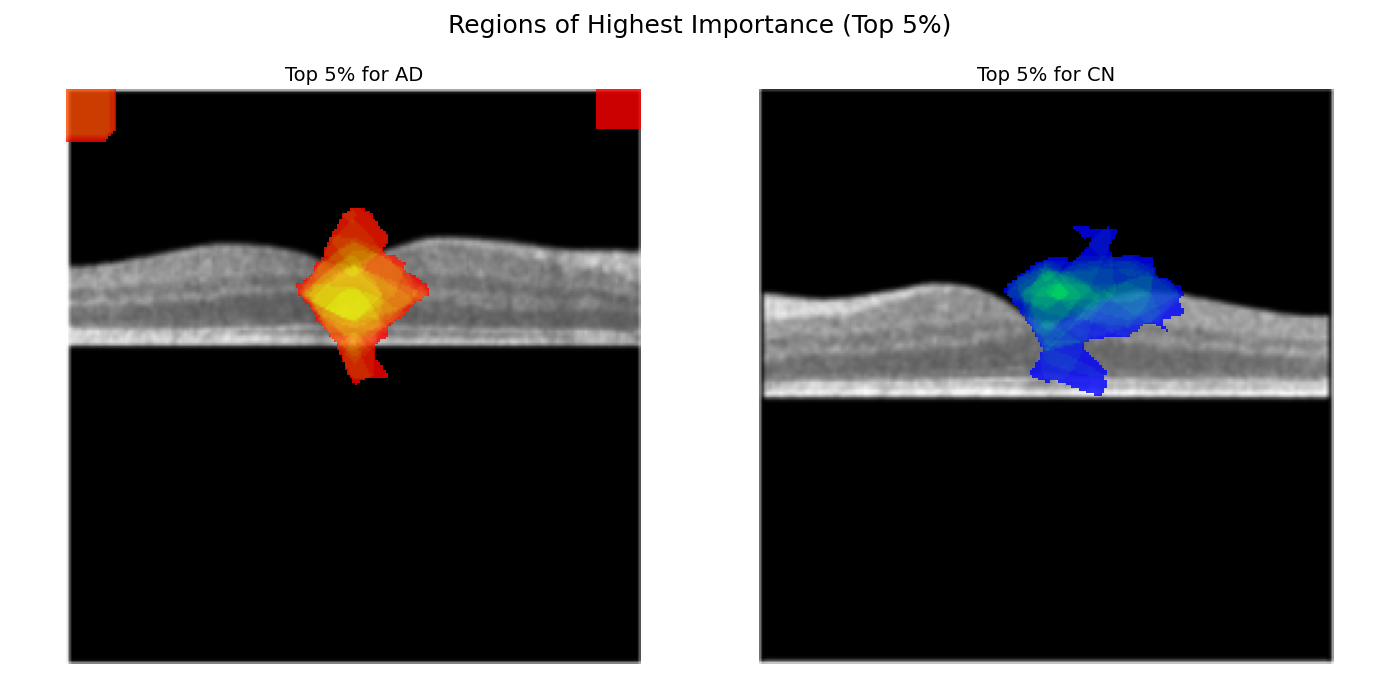}
        \caption{Aggregated Top 5\% Saliency Regions}
        \label{fig:exp_regions}
    \end{subfigure}
    
    \vspace{3mm} % A little vertical space for separation

    % --- Subfigure (b): Individual Examples ---
    \begin{subfigure}[b]{0.95\columnwidth}
        \centering
        \includegraphics[width=\textwidth]{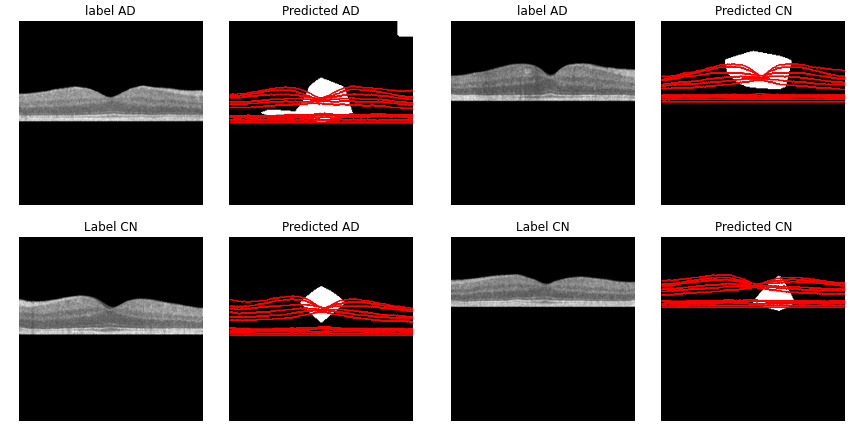}
        \caption{Saliency Maps for Individual Test Samples}
        \label{fig:exp_examples}
    \end{subfigure}

    % --- The MAIN Caption for the whole figure ---
    \caption{Model interpretability analysis. 
    (\subref{fig:exp_regions}) The aggregated top 5\% most salient pixels for the AD (left, red/yellow) and CN (right, blue/green) classes highlight the model's consistent focus on distinct anatomical regions.
    (\subref{fig:exp_examples}) Individual examples show model attention for True Positives (TP), True Negatives (TN), False Negatives (FN), and False Positives (FP). \textit{Images used with permission from the UK Biobank under Application Number 82266.}}
    \label{fig:main_explainability}
\end{figure}

To investigate the  classification decision of the model and ensure that it learns clinically relevant features, we generated and analyzed saliency maps using GradCAM. The results are shown in Fig.~\ref{fig:main_explainability}. First, we aggregated the top 5\% most salient pixels from all test images for each class to identify the most consistently important regions for classification (Fig. ~\ref{fig:main_explainability}(\subref{fig:exp_regions})). Furthermore, we analyzed individual cases to connect the model's attention to its performance on specific examples (Fig. ~\ref{fig:main_explainability}(\subref{fig:exp_examples})). 

The saliency analysis in Table \ref{tab:saliency_overlap} shows that the model gave minimal attention to the RNFL (Filling Ratio $<$ 6\% for AD) and instead focused on the central macular region (34.9\% Filling Ratio for AD), especially the BMEIS and IS/OSJ layers.

%\begin{figure*}
%\centering
%\includegraphics[width=0.8\textwidth]{CAM_Resnet.png}
%\caption{GradCAM and layer-based explainability results for ResNET}
%\label{fig:GradCamRes}       
%\end{figure*}

% =================================================================
% BEGIN COMPACT EXPLAINABILITY FIGURE
% =================================================================

% =================================================================
% END COMPACT EXPLAINABILITY FIGURE
% =================================================================

% =================================================================
% BEGIN QUANTITATIVE OVERLAP TABLE
% =================================================================
\begin{table}[ht!]
    \centering
    \caption{Quantitative Overlap of Saliency Maps with Retinal Layers}
    \label{tab:saliency_overlap}
    % --- Begin group for local settings ---
    \begingroup
    \setlength{\tabcolsep}{5pt} % Aggressively reduce column spacing
    \small % Use a smaller font
    \begin{tabular}{l cc cc cc}
        \toprule
        \multirow{2}{*}{\textbf{Layer}} & \multicolumn{2}{c}{\textbf{IoU}} & \multicolumn{2}{c}{\textbf{Dice}} & \multicolumn{2}{c}{\textbf{Fill (\%)}} \\
        \cmidrule(lr){2-3} \cmidrule(lr){4-5} \cmidrule(lr){6-7}
        & \textbf{CN} & \textbf{AD} & \textbf{CN} & \textbf{AD} & \textbf{CN} & \textbf{AD} \\
        \midrule
        RNFL-GCL & .020 & .013 & .039 & .025 & 10.6 & 5.6 \\
        GCL-IPL & .038 & .017 & .070 & .033 & 15.8 & 7.4 \\
        IPL-INL & .048 & .028 & .089 & .054 & 15.7 & 8.6 \\
        INL-OPL & .052 & .021 & .097 & .041 & 22.7 & 10.7 \\
        OPL-HFL & .067 & .034 & .124 & .065 & 23.7 & 12.6 \\
        BMEIS & \textbf{.164} & \textbf{.103} & \textbf{.273} & \textbf{.177} & \textbf{26.3} & \textbf{18.6} \\
        IS/OSJ & .022 & .015 & .042 & .029 & 16.8 & 13.4 \\
        IB\_OPR & .022 & .017 & .042 & .033 & 15.7 & 14.9 \\
        IB\_RPE & .019 & .015 & .036 & .029 & 11.6 & 10.9 \\
        OB\_RPE & .016 & .015 & .030 & .028 & 9.5 & 9.9 \\
        \midrule
        Macula & .249 & .192 & .379 & .304 & 41.3 & 34.9 \\
        \bottomrule
    \end{tabular}
    \endgroup
    % --- End group ---
    
    \vspace{1mm}
    \begin{minipage}{\columnwidth} % IMPORTANT: Use \columnwidth
        \small 
        \textit{Note:} "Fill (\%)" is the Filling Ratio. The filling Ratio is defined as $|\text{Saliency} \cap \text{Layer}| / |\text{Layer}|$. The bold values indicate the highest overlap for each class within the specific layers.
    \end{minipage}
\end{table}
% =================================================================
% END QUANTITATIVE OVERLAP TABLE
% =================================================================

%%%%%%%%% <<SERDAR ENDS>> %%%%%%%%%%%%%%%

%RETFound-S (vit-L)       & 0.554 & 0.462 & 0.515 & 0.600 &0.429 &0.552 &0.090  & 0.22    \\ %20250822_184417

%% file: discussion.tex
\section{DISCUSSION}
\label {sec:Discussion}

This study presents the first end-to-end deep learning framework for predicting Alzheimer’s disease (AD) from UK Biobank Optical Coherence Tomography (OCT) B-scans up to four years before diagnosis. Our best-performing model, ResNet-34, achieved a mean AUC (mAUC) of $0.624 \pm 0.060$ on a rigorously selected age, sex, and instance-matched AD and control sample. 

Earlier studies using data from the UK Biobank predicted early AD from fundus images. Tian et al \cite{Tian2021}. achieved an accuracy of 0.824 using fundus images with vessel segmentation.  Wisely et al. \cite{Wisely2022} obtained an AUC of 0.625 using only OCTA images and 0.681 with GC-IPL projection maps. However, when they used quantitative data, including age and sex, their model achieved an AUC of 0.96. When all the information (image and quantitative) was combined, the AUC was 0.809. Chua et al \cite{Chua2025}. obtained an AUC of 0.82 when using a GC-IPL projection map dataset; however, their performance was reduced to 0.76 when trained with age-matched subjects. These results show that age and sex can be strong confounding factors in AD prediction models. There is no OCTA image dataset in the UK Biobank; therefore, these studies \cite{Wisely2022,Chua2025} used private OCTA image datasets. Our goal was to study only the structural retinal changes related to AD. As a result, our model performance was moderate, but it likely reflects AD-related structural features in the retina more directly, as it was not influenced by demographic confounders. 

When comparing the architectures, ResNet-34 outperformed both VGG-11 and OCT-pretrained transformer models. VGG-11 achieved a mAUC of 0.581 $\pm$ 0.017, which was not significantly different from that of ResNet-34 after correction. RETFound-C reached an mAUC of 0.540 ± 0.037, which was not significantly different. In contrast, RETFound-S performed considerably worse, with an mAUC of 0.459 ± 0.068, and the difference relative to ResNet-34 remained statistically significant even after correction for multiple comparisons. These findings suggest that in low-sample settings, convolutional networks may provide more stable representations than transformer-based models, despite being pre-trained on OCT data. To further test robustness, we repeated the analysis using the same AD cohort and a randomly matched control group. Performance was preserved (AUC = 0.652 ± 0.058), indicating that the results were stable against variations in the control selection.

Ablation experiments further emphasized the importance of the three-channel input strategy. When ResNet was trained using only masked images, only replicated grayscale OCT, or only retinal layer contours, performance dropped substantially compared to the full multichannel input. This demonstrates that the combination of raw OCT intensity, layer-specific masking, and anatomical boundary (contour) information provides the most informative feature representations. 

Explainability analysis showed that the model allocated very little attention to the RNFL (Filling Ratio $<$6\% for AD), despite its frequent use as a biomarker in previous studies. Instead, attention was concentrated on the central macular region (34.9\% for AD), with the highest overlap observed in the BMEIS and IS/OSJ layers. This indicates that the network relies on features within the photoreceptor and RPE complex, highlighting regions that may act as early biomarkers of AD. These results further support the potential of retinal OCT combined with AI for noninvasive and scalable early AD risk prediction. 

Despite these promising results, this study has several limitations. The dataset size, particularly in the 4-year diagnostic group, constrained the statistical analyses and may have limited the generalizability of the findings to broader populations. One of the main contributing factors to the modest prediction accuracy was the study's dependency on a single dataset, which limited the model’s ability to generalize and increased the risk of overfitting. Although our nested cross-validation framework was designed to mitigate overfitting and provide a robust internal estimate of model performance, external validation remains a critical next step. However, to the best of our knowledge, no publicly available OCT datasets exist that include both retinal imaging and longitudinal Alzheimer's disease outcomes, making such validation infeasible. Future collaborations with clinical centers or research initiatives will be essential to acquire independent cohorts for further validation. Establishing shared datasets and benchmarks in this domain would greatly enhance reproducibility and comparability across studies and facilitate the translation of OCT-based biomarkers into clinical use. Additionally, while nested cross-validation mitigates overfitting, further improvements could be achieved through ensembling, multimodal integration (e.g., OCT angiography, cognitive testing), or temporal modeling of longitudinal scans. Finally, clinical validation against gold-standard biomarkers (e.g., amyloid PET and CSF tau) is necessary to establish OCT’s role in preclinical AD screening.

%% file: conclusion.tex
\section {Conclusion}
\label {sec:Conclusion}

This study presents the first deep learning framework applied to UK Biobank retinal OCT data for early prediction of Alzheimer’s disease up to four years before diagnosis. We proposed an anatomically guided preprocessing pipeline with multichannel OCT input, hybrid augmentation, and nested cross-validation. ResNet-34 achieved the best performance, with a mean AUC of 0.624, and robustness was confirmed with a re-matched control dataset. Ablation experiments showed that combining raw, masked, and contour inputs improved performance, while saliency maps highlighted the macular BMEIS and IS/OSJ layers. 
Our findings provide a reproducible baseline for OCT-based AD prediction, highlight the challenges of detecting subtle retinal biomarkers years before AD diagnosis, and point to the need for larger datasets and multimodal approaches.

%This study provides the first OCT B-scan–based deep learning in AD using  datasets with carefully matched UK Biobank datasets; therefore, the results could set  a benchmark  and a reproducible pipeline for future research.